%% file: main.tex
\documentclass{llncs} 
\usepackage{amsmath}
\usepackage{todonotes}
\usepackage{lscape}
\usepackage{amssymb}
\usepackage{amsfonts}
\usepackage{stmaryrd}
\usepackage{array}
\input{mathmacros.tex}

\title{Large Random Forests: \\
Optimisation for Rapid Evaluation}
\author{
  Frederik Gossen \and
  Bernhard Steffen}
\institute{
  Chair for Programming Systems, TU Dortmund University, Germany \\
  \email{\{frederik.gossen,bernhard.steffen\}@tu-dortmund.de}}
\begin{document}
\maketitle
\begin{abstract}
  Random Forests are one of the most popular classifiers in machine learning.
  The larger they are, the more precise is the outcome of their predictions. 
  However, this comes at a cost: their running time for classification grows linearly with the number of trees, i.e. the size of the forest. 
  In this paper, we propose a method to aggregate large Random Forests into a single, semantically equivalent decision diagram. 
  Our experiments on various popular datasets show speed-ups of several orders of magnitude, while, at the same time, also significantly reducing the size of the required data structure.

\end{abstract}
\section{Introduction}
Random Forests are one of the most widely known classifiers in machine learning~\cite{Ho1995,Breiman2001}.
The method is easy to understand, implement, and at the same time achieves impressive classification accuracies in many applications.
Compared to other methods such as neural networks, Random Forests are fast to train and they are clearly more suitable for smaller datasets. 
In contrast to a single decision tree, Random Forests, a collection of many trees, do not overfit as easily on a dataset and their variance decreases with their size.
On the other hand, their running time for classification linearly grows with this size, which is critical as forests may well consist of thousands of trees -- a problem especially for applications with a high throughput~\cite{facebook}. 

In this paper, we present an optimisation method that is based on radical aggregation: 
Forests are transformed into a single decision diagram in a semantics-preserving fashion, which, in particular, also preserves the learner's variance and accuracy. 
Being a post-process, also the ease of Random Forest training is maintained. 
The great advantage of these decision diagrams is their absence of redundancy: during classification every predicate is considered at most once.

Our experiments with popular data sets from the UCI Machine Learning Repository~\cite{uci-repo} showed performance gains of several order of magnitude (cf. Fig.~\ref{fig:running-time} and Table~\ref{tab:running-time}). 
A potential problem is only an explosion in size which can, in principle, be exponential for decision diagrams. 
However, this problem did not arise in our experiments. 
On the contrary, we even observed drastic size reductions (cf. Fig.~\ref{fig:size} and Table~\ref{tab:size}).

Key to our approach are Algebraic Decision Diagrams (ADDs)~\cite{Bahar1993}. 
Their algebraic structure supports aggregation, abstraction, and reduction operations that we utilise to optimise both, the running time needed for classification and the size of the final data structure. 
When combined with a reduction that exploits the unfeasibility of paths in decision diagrams, this leads to impressive performance gains in our experiments:
\begin{itemize}
  \item Using basic algebraic operations, such as concatenation and addition, allows to aggregate a Random Forest into a single ADD. 
  \item Abstracting results to the essence, in this case the outcome of a majority vote, leads to quite dramatic reductions, both of classification times, as well as of size requirements.
  \item Subsequent elimination of unfeasible paths (path with contradicting predicates) finally achieves our goal: high performance gains (cf. Fig.~\ref{fig:running-time} and Table~\ref{tab:running-time}) and, additionally, significant reductions of space requirements (cf. Fig.~\ref{fig:size} and Table~\ref{tab:size}).
\end{itemize}

Please note that these results are achieved for a widely used classifier and for utterly standard datasets. 
This indicates the generality of our approach to aggregate Random Forests with thousands of trees into a compact decision diagram for rapid classification -- faster by multiple orders of magnitude. 

\medskip

\textbf{Related work:} Runtime performance of Random Forests has been addressed, e.g., via optimising code generation with moderate success~\cite{treelite,weka,facebook,Browne2019}, and, with a greater performance impact, via model simplification which, however, changed the semantics~\cite{Kargupta2004}. 
Others applied semantic aggregation~\cite{Mulvaney2003,Kargupta2004,Giabbanelli2015,Bonfietti2015} to Random Forests, however, without explicitly addressing the runtime performance, while the authors in~\cite{Peterson2009,Painsky2019} were focusing solely on the memory footprint, all with moderate success.

The only paper on Random Forests we know of that uses decision diagrams similar our ADDs is \cite{Nakahara2017}. 
However, they use these diagrams only to compact the individual tree and not to aggregate an entire forest. 
In fact, the reported speedup by a factor of up to $61$ seems more to rely on technical and even hardware details than on the use of decision diagrams. 
In contrast, our approach focuses on the decision diagram-based holistic aggregation of entire Random Forests, which, due to its globality, has a much greater impact. 
In fact, we obtain speed-ups already at the hardware-independent level that are orders of magnitude higher than in~\cite{Nakahara2017}.

\medskip

After a short introduction to Random Forests in Section~\ref{sec:random-forest}, we present our approach to their aggregation in Section~\ref{sec:label-vector}, which is subsequently refined in two steps: by compositional and non-compositional abstraction in Section~\ref{sec:abstraction}, and by the elimination of redundant predicates from the decision diagrams in Section~\ref{sec:unsatisfiable-path-elimination}. 
The impact on the classification time and size of the new decision diagrams is evaluated in Section~\ref{sec:evaluation}. 
The paper closes with conclusions and direction to future work in Section~\ref{sec:conclusion}.

\section{Random Forests}
\label{sec:random-forest}
Random Forests is one of the most widely known classifiers in machine learning.
The algorithm is relatively simple and yields good results for many real-world applications.
Its decision model generalises a training dataset that holds examples of input data labelled with the desired output, also called \emph{class}.
As its name suggests, a Random Forest consists of some number of decision trees.
Each of these trees is itself a classifier that was learned from a random sample of the training dataset.
Consequently, all trees are different in structure, they represent different decision functions, and can yield different decisions for the very same input data.

To apply a Random Forest to previously unseen input data, every decision tree is evaluated separately:
Tracing the trees from their root down to one of the leaves yields one decision per tree, i.e. the predicted class. 
The overall decision of the Random Forest is then derived as the most frequently chosen class, an aggregation commonly referred to as \emph{majority vote}. 
Key advantage of this approach is the, compared to single decision trees, reduced variance.
A detailed introduction to Random Forests, decision trees, and their learning procedures can be found in~\cite{Ho1995,Breiman2001,Quinlan1986}.

In this paper, we use Weka~\cite{weka} as our reference implementation of Random Forests.
However, our approach does not depend on implementation details and can be easily adapted to other implementations.

\begin{figure*}[t]
  \centering
  \includegraphics[width=.9\textwidth]{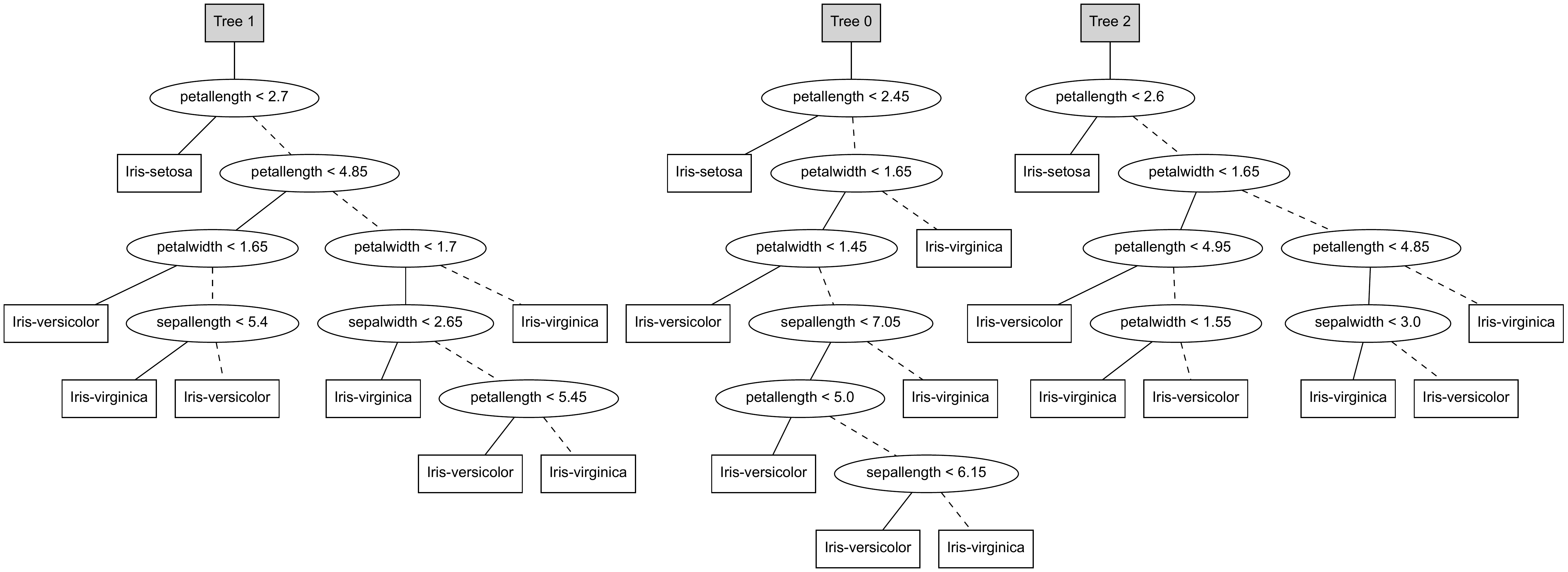}
  \caption{Random Forest learned from the Iris dataset~\cite{Fisher1936}.}
  \label{fig:random-forest}
\end{figure*}
Figure~\ref{fig:random-forest} shows a small Random Forests that was learned from the popular Iris dataset~\cite{Fisher1936}. 
The dataset lists dimensions of Iris flowers' sepals and petals for three different species. 
Using this forest to decide the species on the basis of given measurements requires to first evaluate the three trees individually and to subsequently determine the majority vote. 
This effort clearly grows linearly with the size of the forest. 
In the following we use this example to illustrate our approach of forest aggregation and its great effect on the required evaluation effort.

\section{Rapid Evaluation of Random Forests}
\label{sec:label-vector}
To evaluate a Random Forest, we are forced to evaluate every single tree separately.
While our illustrative example consists of only three trees there is essentially no limit to the number of trees in a forest.
In fact, increasing its size can only improve the classifier and will, in contrast to other classifiers, not lead to overfitting~\cite{Breiman2001}.
However, with a growing number of trees comes increasing computational cost for its construction and, more importantly, for its classification process.
In this paper we focus on the costs associated with the model's classification process (the classification time), while accepting additional effort for their construction.
This focus reflects the fact that constructed decision structures, once deployed, are often meant to be used by millions of users in parallel. 

Key idea behind our approach is to partially evaluate the Random Forests at construction time which, we will see, has an enormous impact on the classification performance and the corresponding space requirements. 
This is, not the least, due to the fact that the individual trees of a Random Forest typically share some similarities. 
E.g., in our accompanying Iris flower example (cf. Fig.~\ref{fig:random-forest}) the predicate $petalwidth~<~1.65$ is used in all three trees. 
This can easily lead to cases where the same predicate is evaluated many times in the classification process. 
The partial evaluation proposed in this paper transforms Random Forests into decision structures where such redundancies are totally eliminated.  

An adequate data structure to achieve this goal for binary decisions are Binary Decision Diagrams~\cite{Bryant1986,Akers1978,Lee1959} (BDDs):
For a given predicate ordering, they constitute a normal form where each predicate is evaluated at most once, and only if required to determine the final outcome.

Algebraic Decision Diagrams (ADDs)~\cite{Bahar1993} generalise BDDs to capture functions of the type $\mathbb{B}^\mathcal{P} \rightarrow \mathcal{C}^n$ which are exactly what we need to specify the semantics of Random Forests for a classification domain $\mathcal{C}$. 
Moreover, in analogy to BDDs, which inherit the algebraic structure of their co-domain $\mathbb{B}$, ADDs also inherit the algebraic structure of their co-domains if available. 

We exploit this property during the partial evaluation of Random Forest by considering two algebraic co-domains, the class word co-domain $\mathcal{C}*$ (cf. Sec.~\ref{subsec:label-vector-monoid}) and the class vector co-domain (cf. Sec.~\ref{subsec:abstraction-label-multiset}).
The aggregation to achieve the corresponding optimised decision structures is then a straightforward consequence of the used ADD technology.

\subsection{Algebraic Structure for Random Forest Results}
\label{subsec:label-vector-monoid}
Let us put aside the evaluation procedure for now and focus only on its outcome, i.e. the final result.
The unprocessed result of a Random Forest evaluation is an ordered sequence with one decision per tree.
This sequence preserves all information and remains independent of any particular aggregation method, e.g. \emph{majority vote}.

To formalise this, let $\mathcal{C}$ be the set of classes, e.g. the three Iris flower species.
An individual tree's decision is one class $c \in \mathcal{C}$, a Random Forest's decision is a word over these classes $\mathbf{c} \in \mathcal{C}^*$.
Note that we can describe the results of any Random Forests this way, no matter its size.
In particular, we can represent the decision made by the empty Random Forest with the empty word $\epsilon$ and the results of a single decision tree with a word of length one.
Moreover, this representation naturally allows for composition: we can simply concatenate the results of two distinct Random Forests, maintaining a one-to-one association between the word's symbols, i.e. the classes, and the corresponding tree in the forest.

With that in mind, we can define the algebraic structure \emph{class words} as a string monoid to represent the results of any Random Forest:
\begin{align*}
  W &:= (\mathcal{C}^*, \circ, \epsilon) \textrm{.}
\end{align*}
The classes $\mathcal{C}$ form its alphabet, concatenation $\circ$ is its associative join operation, and the empty word $\epsilon$ serves as a neutral element of the monoid.

% Recall the example of an Iris flower with a petal length of 2.5cm, a width of 1.5cm, a sepal length of 6cm, and a width of 2.7cm.
% Again, while the first decision tree predicts the species \emph{Iris versicolor} the two others predict \emph{Iris setosa} to be the more likely species.
% With our algebraic structure $V$, this outcome is simply encoded in the vector
% %
% \begin{align*}
%   \mathbf{c} &= (Iris\mhyphen{}versicolor, Iris\mhyphen{}setosa, Iris\mhyphen{}setosa) \fullstop
% \end{align*}
% 
\subsection{Semantics-preserving Transformation}
\label{subsec:label-vector-transform}
With \emph{class words}, we have discussed a means to represent only the outcome of a Random Forest classification.
To transform the entire Random Forest to an Algebraic Decision Diagram (ADD), however, we need to transform not only the results but also the decision structure.
To this aim, we must guarantee the unique properties of decision diagrams: 
\begin{itemize}
  \item they enforce an order of predicates along all paths, and 
  \item they are directed acyclic graphs that share common substructures where possible.
\end{itemize}

With the compositionality of the algebraic structure $W$ and the corresponding ADDs $\mathcal{D}_W$, we can transform any Random Forest incrementally.
Starting with the empty Random Forest, we consider one tree after the other, aggregating a growing sequence of decision trees until the entire forest is entailed in the new decision diagram.
We will first find a semantically equivalent decision diagram for the empty Random Forest, the neutral element of this aggregation procedure.
Subsequently, we will describe a semantics-preserving transformation for single decision trees and a join operation to incorporate these decision diagrams into the overall aggregation.

No matter the input it was given, the empty Random Forest with $0$ trees can only result in one outcome: the empty word $\epsilon$.
Hence, it resembles the constant function, also denoted $\epsilon$ for brevity, that is semantically equivalent to the constant decision diagram with $\epsilon$ as its only terminal node.
This diagram forms the neutral element of our aggregation procedure.
To transform a single decision tree, we can build upon the well-known ADD construction operation $ite$.
For a predicate $p$ and two decision diagrams $f$ and $g$, $ite(p, f, g)$ constructs the diagram that evaluates to $f$ if $p$ holds and to $g$ otherwise.
We derive the decision diagram recursively along the tree structure, effectively delegating the entire model transformation to well-known and efficient algorithms in a service-oriented fashion.
The algorithm implementing $ite$ ensures a strict predicate order and automatically shares substructures where possible.
In fact, the resulting decision diagram is a canonical representation of the function for a given predicate order.
Formally, this defines a function $d_W: \mathcal{T} \rightarrow \mathcal{D}_W$ mapping decision trees to decision diagrams over \emph{class words}:
\begin{align*}
  d_W(t) &:=
  \begin{cases}
    t_{val} & \textrm{if } t \textrm{ is leaf,} \\
    ite(t_{pred}, d_W(t_{then}), d_W(t_{else})) & \textrm{otherwise.}
  \end{cases}
\end{align*}

Having transformed every decision tree individually leaves us with the task to compose the resulting sequence of decision diagrams.
% On one hand, these diagrams are a compact and canonical representation of functions, on the other hand, they allow for efficient operations on them.
This is where the above-mentioned characteristic properties of ADDs come into play: they inherit the algebraic structure of their co-domains, and they even come with efficient algorithms for computing the required operations. 
In this case, this concerns only the concatenation $\circ$ of words over the carrier set $\mathcal{C}^*$.
To ease readability, we denote also this terminal-wise concatenation of decision diagrams with the same symbol~$\circ$.

The desired decision diagram, aggregating an entire sequence of decision trees $t_0, t_1, \dots, t_n$, can now simply be defined as follows:
\begin{align*}
  d_W(t_0, t_1, \dots, t_{n - 1}) &:= d_W(t_0) \circ d_W(t_1) \circ \dots \circ d_W(t_{n - 1}) \textrm{.}
\end{align*}

% Because of the associativity of $\circ$ and the neutral element $\epsilon$, we can lazily construct decision diagrams from decision trees one after the other.
% We aggregate only one decision diagram and incorporate one tree after the other until we have reached a semantically equivalent representation for the entire Random Forest.
% 
\begin{figure*}[t]
  \centering
  \includegraphics[width=.9\textwidth]{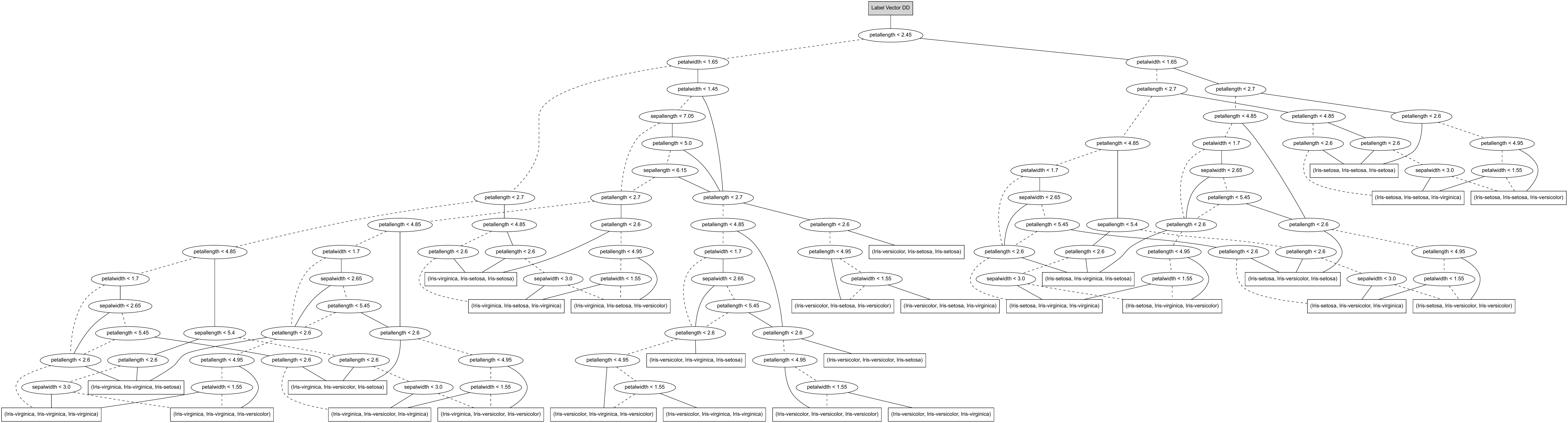}
  \caption{Partially evaluated Random Forest.}
  \label{fig:label-vector-dd}
\end{figure*}
Figure~\ref{fig:label-vector-dd} shows the aggregation of our exemplary Random Forest (cf. Fig.~\ref{fig:random-forest}).
% Its semantics are preserved while redundant evaluation of predicates is eliminated.
Already, for this extremely small example, the average running time for classification is reduced. 
Its true impact, however, becomes apparent only with increasing forest size (cf. Sec.~\ref{sec:evaluation}).

\section{Partial Evaluation through Abstraction}
\label{sec:abstraction}
\emph{Class words} faithfully represent the information about the decisions of each individual tree in the forest.
This is far more than necessary for determining the final decision of the forest, the majority vote. 
In this section we exploit this leeway as follows:
\begin{itemize}
  \item We first abstract from the information which tree was responsible for which decision by moving from the domain of class words to the domain $\mathbb{N}^{|\mathcal{C}|}$ of \emph{class vectors} which simply record the frequency with which each class has been proposed.
  $(\mathbb{N}^{|\mathcal{C}|,+)}$, where addition $+$ is defined component-wise, is again a monoid. 
  The structure can again be lifted to the ADD-level in order to guarantee the compositional aggregation of forest as illustrated in the previous section for the class word monoid.
  \item Subsequently, we abstract the aggregated result to a decision diagram which only reflects the majority vote.
  This abstraction is not compositional and can therefore only be applied at the very end of the aggregation process as an additional optimisation.
\end{itemize}
The following subsections are devoted to these two steps, respectively.

\subsection{From Class Words to Class Frequencies}
\label{subsec:abstraction-label-multiset}
As mentioned above, the precise knowledge about the individual decisions of the trees in a Random Forest is
unnecessary for determining the final decision of a Random Forest. 
The knowledge about the frequency with which each class has been proposed suffices. 
This information can elegantly be represented as \emph{class vectors}, where each component represents one class and its value the frequency with with the class was chosen. 
Formally, the domain of \emph{class vectors} forms a monoid 
\begin{align*}
  V &:= (\mathbb{N}^{|\mathcal{C}|}, +, \mathbf{0}) 
\end{align*}
where addition $+$ is defined component-wise and $\mathbf{0}$ is the neutral element.

Based on this structure we can replay the development of the previous section by replacing $W$ by $V$, $\circ$ by $+$, and $\epsilon$ by $\mathbf{0}$. 
Indexing the class vectors directly with the class labels $c \in \mathcal{C}$ rather than integers this reads as follows:

The required representation of the empty decision is again provided by the neutral element, here the $\mathbf{0}$ vector, and a single class $c \in \mathcal{C}$ can be represented naturally by a vector $\mathbf{i}(c)$ that is $0$ everywhere except for position $c$, where it is $1$. 
Also the construction of the ADDs $\mathcal{D}_V$ with class vectors as their terminal values can simply be transcribed, as can the new transformation function $d_V: \mathcal{T} \rightarrow \mathcal{D}_V$, which differs only in its mapping from the tree's leaves to the new carrier set:
\begin{align*}
  d_V(t) &:=
  \begin{cases}
    \mathbf{i}(t_{val}) & \textrm{if } t \textrm{ is leaf,} \\
    ite(t_{pred}, d_V(t_{then}), d_V(t_{else})) & \textrm{otherwise.}
  \end{cases}
\end{align*}
%
% As before, the transformation relies entirely on the decision diagram construction routine $ite$.
Having adopted the underlying algebraic structure, all operations are seamlessly applicable to the corresponding decision diagrams as well.
In this case, vector summation $+$ is lifted to the new decision diagrams over vectors and we can again easily aggregate the Random Forest incrementally: 
% With $+$ being associative and commutative, we can express the final aggregation of decision trees even more compactly than before as
%
\begin{align*}
  d_V(t_0, t_1, \dots, t_{n - 1}) &:= \sum_{i = 1}^{n - 1} d_V(t_i) \textrm{.}
\end{align*}
The new transformation abstracts from the order of class labels but maintains all the information required to construct and aggregate decision diagrams incrementally.

Abstracting from the order of class labels has two important advantages:
\begin{enumerate}
  \item \textbf{Memory} is saved as many leaf nodes that differed only in the order of class labels are now unified.
  In fact, this effect can ripple up the entire decision diagram, i.e. the structure can partially collapse. 
  Moreover, the vector representation itself also becomes more compact. 
  \item \textbf{Classification time} of the decision diagram is reduced as a result of the partial collapse of the structure:
  Where a predicate was previously needed to differentiate between two class words that differed only in the order of their class labels, this evaluation step becomes redundant. 
  Moreover, the final aggregation step reduces to finding the maximal component of a single class vector.
\end{enumerate}
\begin{figure}[t]
  \centering
  \includegraphics[width=\textwidth]{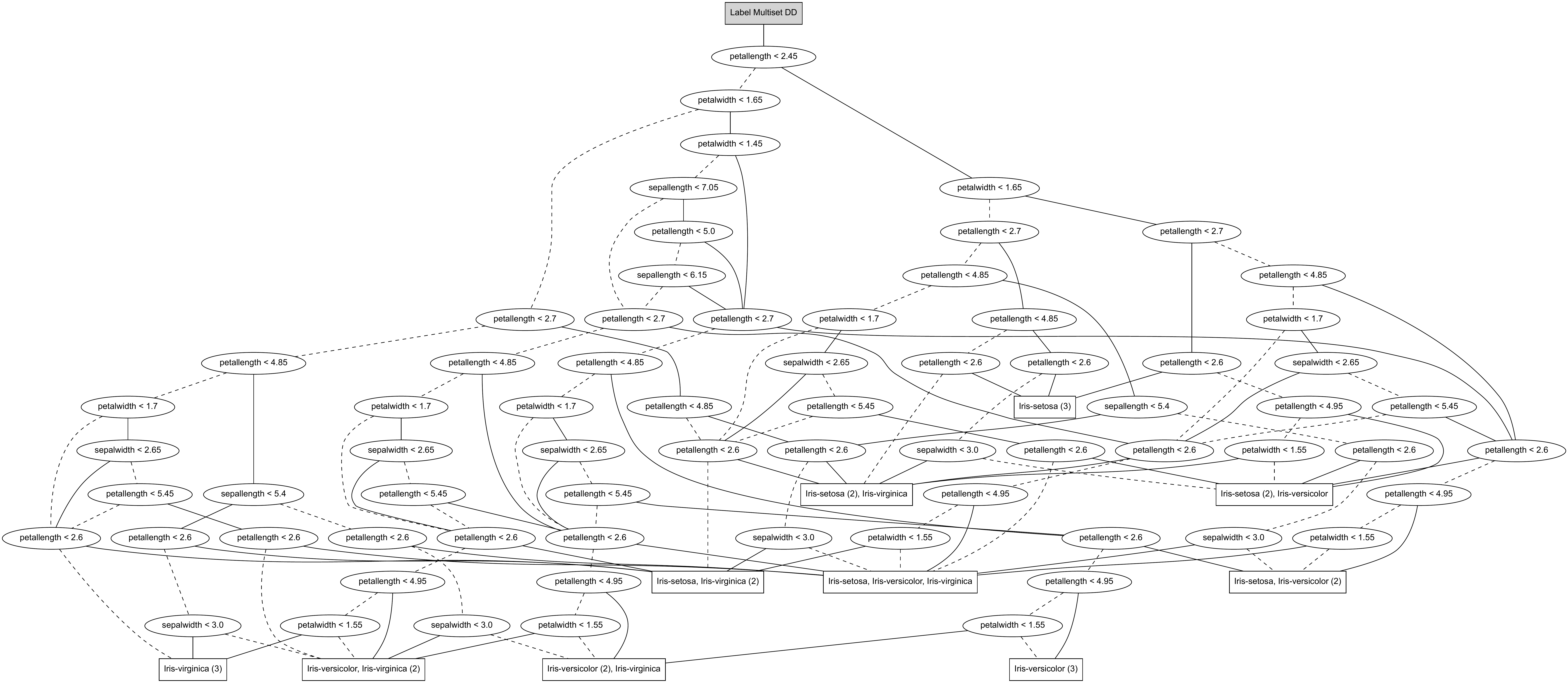}
  \caption{Class vector abstraction of aggregated Random Forest.}
  \label{fig:label-multiset-dd}
\end{figure}
Figure~\ref{fig:label-multiset-dd} shows the result of the class frequency abstraction for our running example.

%\ \\
%\newpage
\subsection{Majority Vote at Compile Time}
\label{subsec:abstraction-most-frequent-label}
As mentioned before, just maintaining the information about the result of the majority votes is not compositional. 
In fact, knowing the result of the majority votes for two Random Forest gives no clue about the majority vote of the combined forest. 
Thus the most frequent class abstraction can only be applied at the very end, after the entire aggregation has been
computed compositionally. 
In fact, the class frequency abstraction provides the most concise compositional abstraction. 
Any further reduction directly leads to potential compositionality violations, or as we say, the class frequency abstraction is \emph{fully abstract} for this scenario.
Thus taking the formerly defined model transformation $d_V$ to iteratively aggregate the trees of a Random Forest is provably the best one can do.
\footnote{A corresponding proof via contraposition is quite straightforward but beyond the scope of this paper.}

The result is a decision diagram $d_V(\mathbf{t}) \in \mathcal{D}_V$ with class vectors in its terminal nodes.
Only the subsequent monadic transformation $mv : \mathcal{D}_V \rightarrow \mathcal{D}_C$ remains to be defined.
With ADDs we can simply define this on the carrier set, i.e. on the class vectors. 
For any class vector $\mathbf{v} \in \mathbb{N}^{|\mathcal{C}|}$ the majority vote is defined as
\begin{align*}
  mv(\mathbf{v}) &:= \argmax_{c \in \mathcal{C}} \mathbf{v}_{c} \textrm{.}
\end{align*}
%
% In case the most frequent class is ambiguous the outcome may be any of them.
Just like binary operations of the algebraic structure $V$ are lifted to Algebraic Decision Diagrams, so can monadic operations~\cite{Bahar1993}.
Note that $mv$ does not project into the same carrier set but rather from one algebraic structure $V$ into another $C$.
However, these transformations can be applied to the corresponding decision diagrams in the very same way.
We can therefore define the final transformations as
\begin{align*}
  d_C(\mathbf{t}) &:= mv(d_V(\mathbf{t})) \textrm{.}
\end{align*}

Post-processing vector decision diagrams in this way has again quite some effect:
Both memory and classification time are reduced for the same reasons as for the class frequency abstraction. 
However, the coarser abstraction leads to stronger reductions. 
Moreover, the aggregation step for determining the final decision of the Random Forest, the majority vote, is no longer necessary.

\begin{figure}[t]
  \centering
  \includegraphics[width=\textwidth]{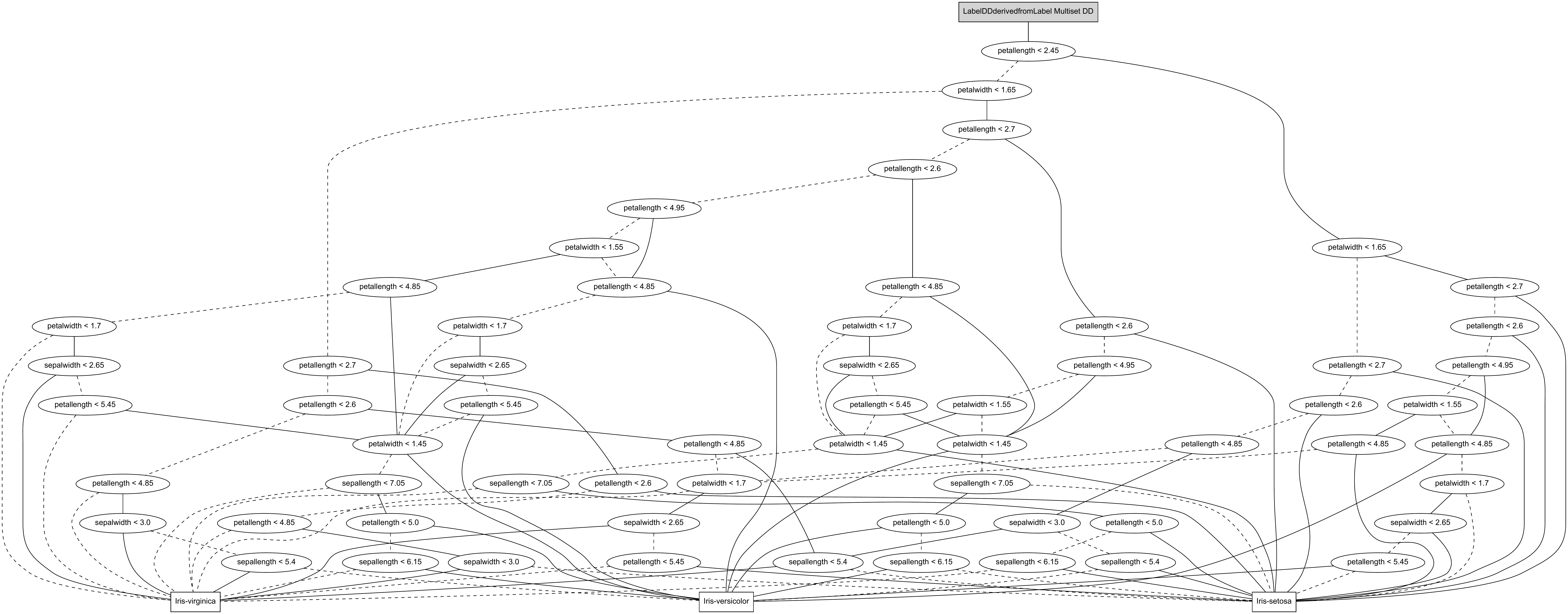}
  \caption{Most frequent label abstraction of aggregated Random Forest (majority vote).}
  \label{fig:label-dd}
\end{figure}
Fig.~\ref{fig:label-dd} shows the result of the most frequent class abstraction for our running example.

\medskip

Our evaluation in Sec.~\ref{sec:evaluation} shows that all the abstractions proposed so far are insufficient to guarantee true scalability, and this despite the fact that they are optimal. 
This is due to the fact that they all are symbolic and do not take the semantics of predicates into account. 
The following section sketches how this shortcoming can be overcome and finally provides us with the scalability results we desired. Please note, however, that the preceding abstractions are necessary to enable the potential of unfeasible path elimination (cf.\ Sec.~\ref{fig:running-time}). 

\section{Unsatisfiable Path Elimination}
\label{sec:unsatisfiable-path-elimination}
\begin{figure}[t]
  \centering
  \includegraphics[width=\textwidth]{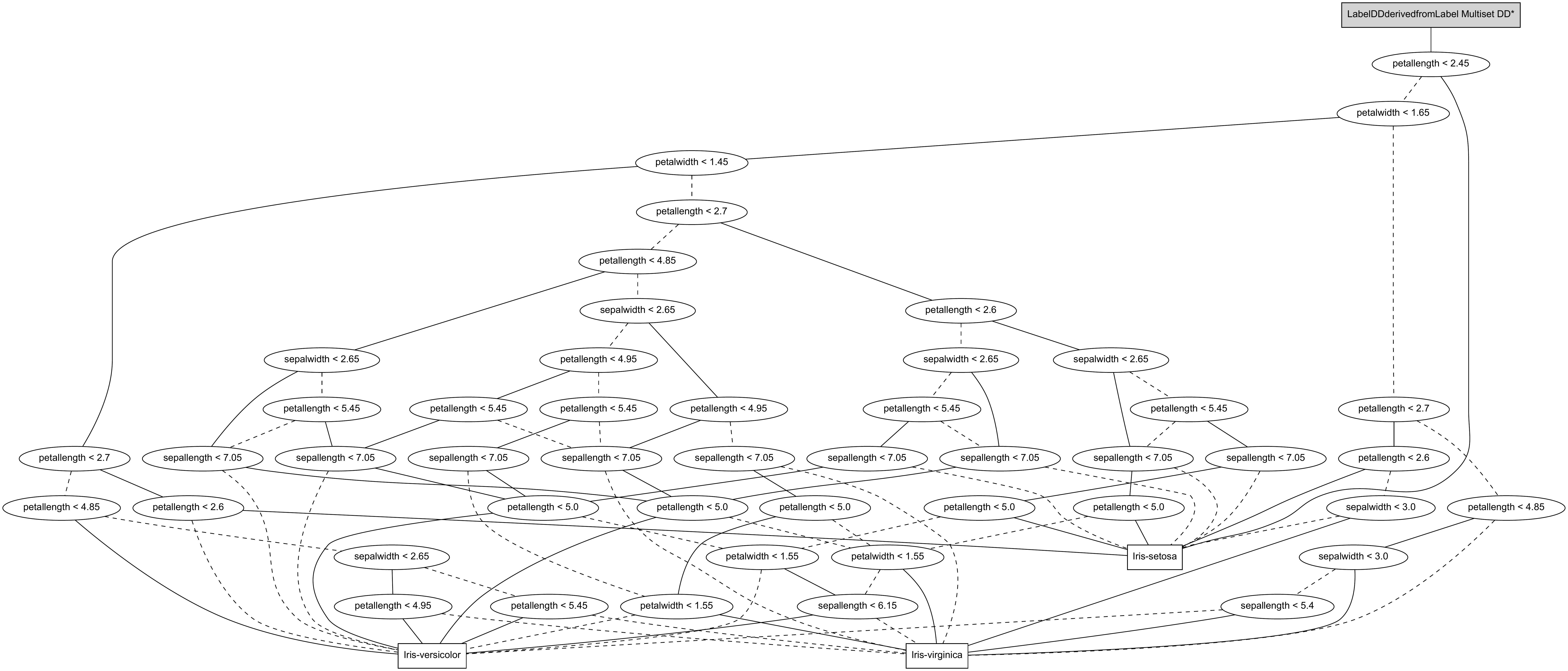}
  \caption{Most frequent label abstraction of aggregated Random Forest (majority vote) without semantically redundant nodes.}
  \label{fig:label-dd-2}
\end{figure}
When aggregating the trees of a Random Forest they all use varying sets of predicates.
In contrast to simple Boolean variables, predicates are not independent on one another, i.e. evaluation of one predicate may yield some degree of knowledge about other predicates.
E.g., the predicate $petallength~<~2.45$ induces knowledge about other predicates that reason about $petallength$:
When the petal length is smaller than $2.45$ it cannot possibly be greater or equal to $2.7$ at the same time.
This is not taken care of by the symbolic treatment of predicates we followed until now.

Unsatisfiable path elimination, as illustrated by the difference between Figure~\ref{fig:label-dd} and 
Figure~\ref{fig:label-dd-2} for our running example, leverages the potential of a semantic treatment of predicates with significant effect:
\begin{itemize}
  \item The \textbf{size} of decision diagrams is drastically reduced (cf. the cases marked with '*' in Fig.~\ref{fig:size}), and even
  \item the \textbf{classification times} further improve because semantically redundant decisions are eliminated (cf. the cases marked with '*' in Fig.~\ref{fig:running-time}).
\end{itemize}
Please note that unfeasible path elimination
\begin{itemize}
  \item depends on previous powerful abstraction: 
  The trees in the original Random Forest have no unfeasible paths by construction. 
  They are introduced in the course of our \emph{symbolic} aggregation, which is insensitive to semantic properties. 
  \item is compositional and can therefore be applied during the stepwise transformation and before the final most frequent label abstraction and at the very end. 
  This avoids that intermediate decision diagrams grow too large which would inhibit the scalability. 
  As can be seen in Figure~\ref{fig:size}, without this effect, our approach would hardly scale to forests beyond the size of 100 trees.
  \item does not support normal forms. 
  Thus our approach may yield different decision diagrams depending on the order of tree aggregation. 
  It is guaranteed, however, that the resulting decision diagrams are minimal.
\end{itemize}

\medskip

Unsatisfiable path elimination is a hard problem in general.\footnote{For the cases considered here it is polynomial, but there are of course theories for which it becomes exponentially hard or even undecidable.} 
Our corresponding implementation uses SMT-solving to eliminate all unsatisfiable paths. 
An in-depth discussion of unsatisfiable path elimination is a topic in its own and beyond the scope of this paper. 

\section{Evaluation}
\label{sec:evaluation}
\begin{figure}[t]
  \centering
  \includegraphics[width=.8\textwidth]{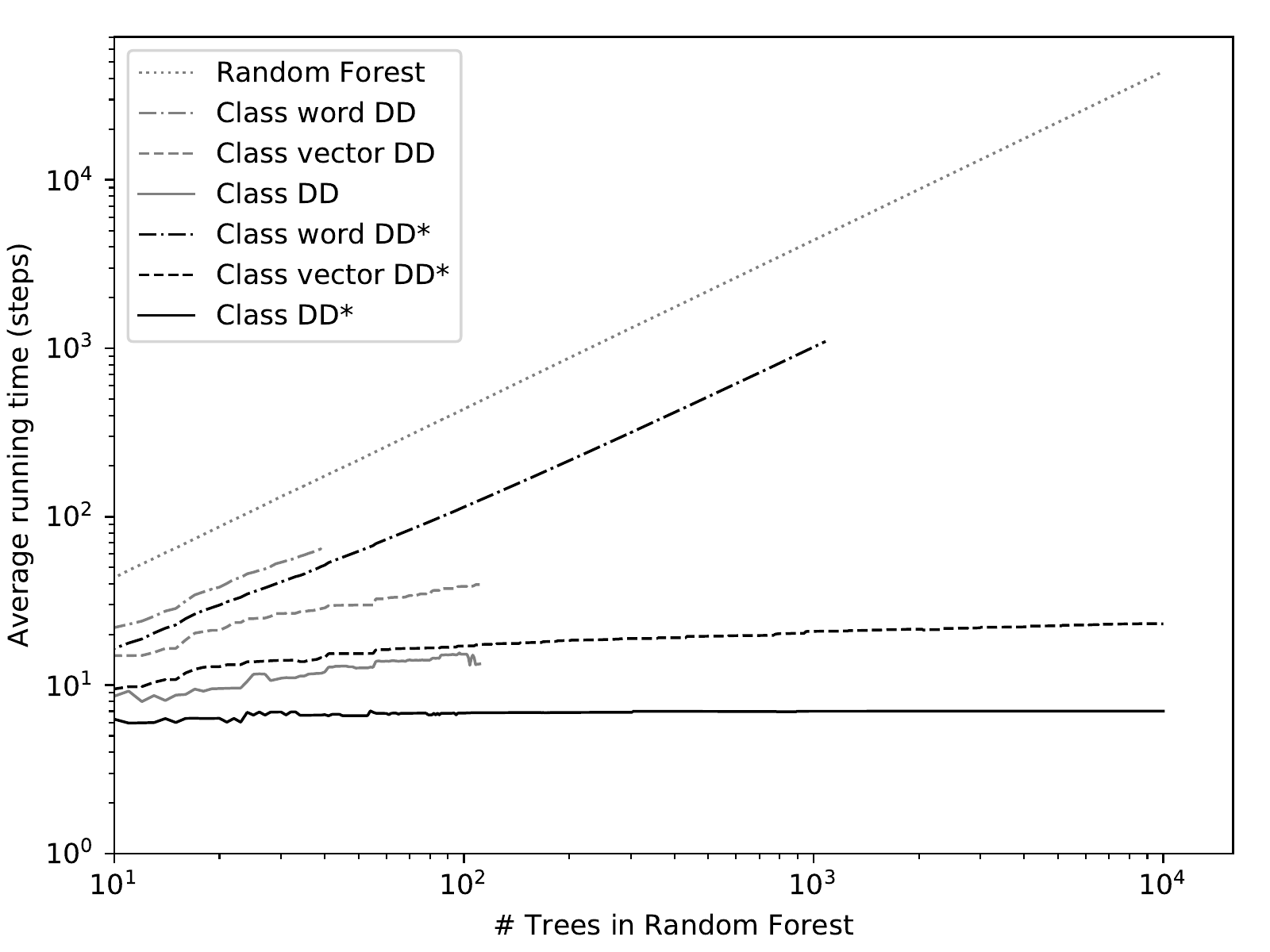}
  \caption{Average running time for classification over all examples in the Iris dataset~\cite{iris-dataset}.}
  \label{fig:running-time}
\end{figure}
\begin{table}[t]
  \centering
  \begin{tabular}{|l|r|r|} 
    \hline
    \textbf{Dataset} & \textbf{Random Forest} & \textbf{Final DD} \\
    \hline
    \hline
    Balance Scale &  80,277.03 &  8.16 (-99.99\%) \\
    Breast Cancer & 130,361.20 & 17.73 (-99.99\%) \\
    Lenses        &  43,883.79 &  3.67 (-99.99\%) \\
    Iris          &  44,043.89 &  7.01 (-99.98\%) \\
    Tic-Tac-Toe   & 107,300.69 & 14.18 (-99.99\%) \\
    Vote          &  69,216.62 &  8.30 (-99.99\%) \\
    % Weather       &  41,888.07 &  5.64 (-99.99\%) \\
    \hline
  \end{tabular}
  \vspace{1em}
  \caption{Running time improvements for classification with Random Forests of size 10.000 for other datasets~\cite{uci-repo}.}
  \label{tab:running-time}
\end{table}
Our three tree accompanying example is useful to explain the concepts but inadequate to illustrate the impact of our radical aggregation technology. 
This section therefore provides a careful quantitative analysis on the basis of a number of popular data sets that illustrates the performance differences between the semantically equivalent representations of the original Random Forest. 

The diagrams in this section show the results concerning quantitative extensions of the Iris flower set which, in the small, also served for our running example. 
The tables summarise the results for other popular data sets to indicate the generality of our approach. 
All the reported classification time and size results were determined as the average over the entire corresponding data sets. 
For the Iris flower example these are 150 records, a number also explaining the quite smooth result graphs.

Our implementation relies on the standard Random Forest implementation in Weka~\cite{weka} and on the ADD implementation of the ADD-Lib~\cite{add-lib,lde,lde-stress}. 
Please note that the considered data sets have been developed with evaluations of this kind in mind, by independent parties, and that we are not using any additional data for our transformation. 
Thus our analysis can be considered unbiased. 

% With regard to the original Random Forests and the different types of decision diagrams, we are primarily interested in the running time needed to evaluate the decision model.
% However, we shall also discuss the price paid for this speedup: the size of the data structure and the cost invested in its construction.
% 
%\subsection{Evaluation Time Reduction}
%\label{subsec:evaluation-running-time}
%
Optimising the classification time is the primary goal of our approach. 
As wall clock time measurements are very sensitive to implementation details and machine profiles, we decided for the, in our eyes more objective measure of step count for our performance analysis. 
As steps we consider here the steps through the corresponding data structures, and in cases where the most frequent class must be computed at runtime, we account one additional step per read. 
For both, the original Random Forest and the word-based decision diagram these are $n$ additional steps and the class vector variant needs $|\mathcal{C}|$ additional steps. 

% This way, the cost to evaluate any predicate as well as the cost to read a leaf or terminal value is $1$.
% We omit the cost associated with a posteriori aggregation of the result, e.g. the cost of majority vote in the Random Forest and two types of decision diagrams.
% Note that this cost can be significant when compared to the cost for tracing down a single decision diagram.
% However, this computation is equally expensive for the original Random Forest, for class vector decision diagrams, and for multiset decision diagrams.
% In contrast to that, no such cost applies to the post-processed class decision diagrams as for these the \emph{majority vote} was pre-evaluated.
% 
Figure~\ref{fig:running-time} shows the average evaluation times of the decision models for Random Forests of up to 10,000 trees.
% The average is taken among all examples from the Iris dataset.
The evaluation time of the original Random Forest grows linearly as expected: every new tree contributes approximately the same running time.
Due to the large number of trees relative to their individual sizes our measurements appear as an almost straight line.
% While there are small variations among the running times of individual trees, the overall running time grows smoothly linear with negligible noise.

Already, the word-based diagrams (cf. Class word DD in Fig.~\ref{fig:running-time}) reduce the classification time significantly in comparison to the original Random Forest. 
This is due to the suppression of redundant predicate evaluations. 
In fact, the overall classification time is dominated by the linearly growing time to compute the most frequent class in each terminal word. 

\begin{figure}[t]
  \centering
  \includegraphics[width=.8\textwidth]{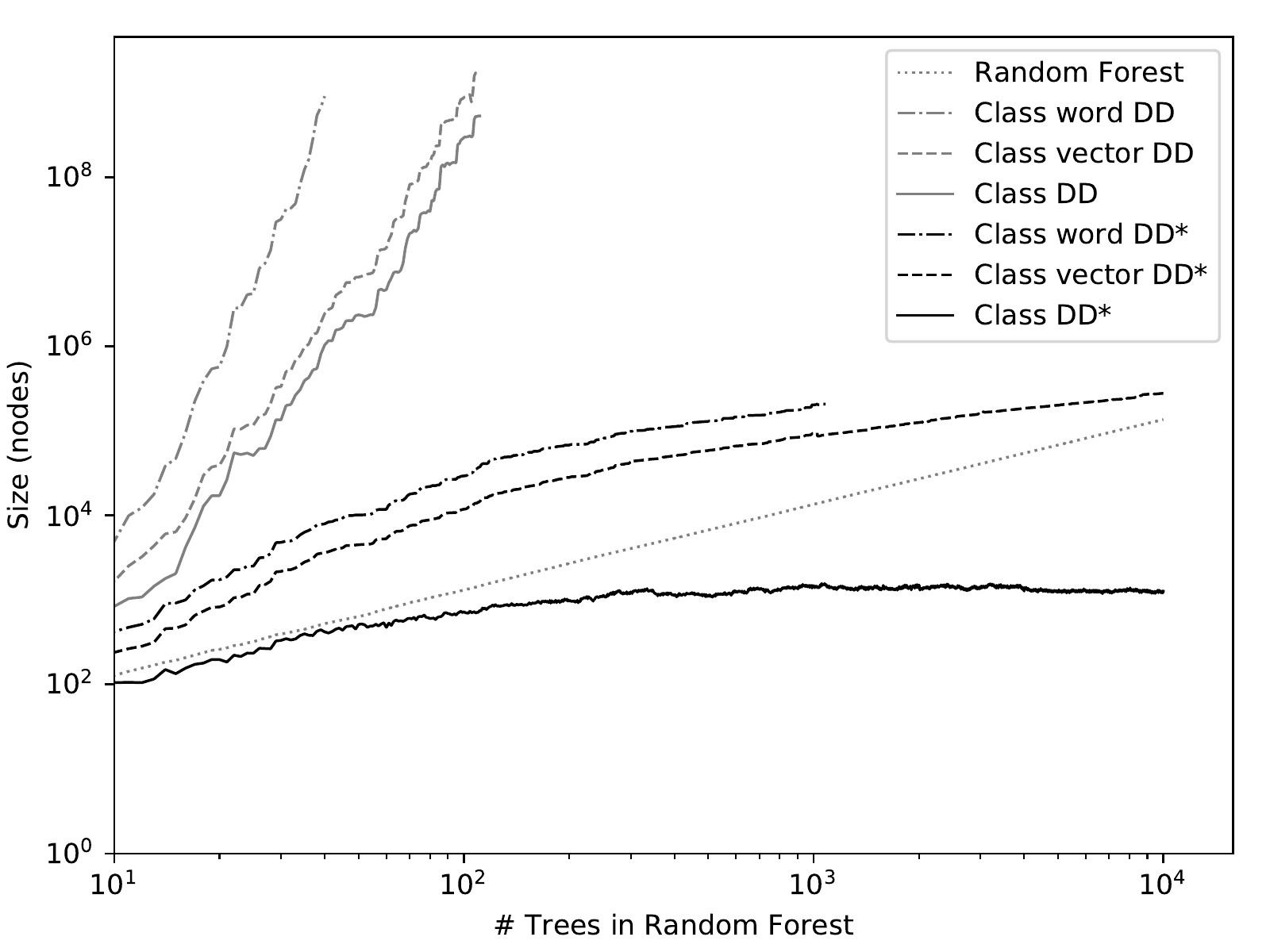}
  \caption{Sizes of the Random Forest and its semantically equivalent decision diagrams.}
  \label{fig:size}
\end{figure}

The reduction to just $|\mathcal{C}|$ terminal nodes of the class vector-based variants has two effects:
\begin{itemize}
  \item A partial collapse of the decision diagram: it is no longer essential which tree proposes which class, unifying all cases where the various classes are equally often proposed. 
  \item Reduction to a constant overhead for the final aggregation step, in this case $|\mathcal{C}|$.
\end{itemize}
The evaluation time reductions are again quite significant, only the space requirement got, like for the word-based variant, out of hand very soon (cf. Fig.~\ref{fig:size}), explaining the cut-off in Fig.~\ref{fig:running-time}.

Whereas the previous two model structures can directly be computed compositionally, the most frequent label abstraction, i.e.  the evaluation of the \emph{majority vote} at compile time, can only be applied at the very end. 
Thus its construction has the same limitation as the class vector variant, and its impact on the size of the corresponding decision model is moderate (cf. Fig.~\ref{fig:size}). 
Its impact on the evaluation time is, however, quite substantial (cf. Fig.~\ref{fig:running-time}): 
Many of the internal decision nodes have become redundant by just focusing on the results of the majority vote.

\begin{table}[t]
  \centering
  \begin{tabular}{|l|r|r|} 
    \hline
    \textbf{Dataset} & \textbf{Random Forest} & \textbf{Final DD} \\
    \hline
    \hline
    Balance Scale & 2,158,330 &   144 (-99.99\%) \\
    Breast Cancer & 5,494,682 & 3,760 (-99.93\%) \\
    Lenses        &   136,986 &    11 (-99.99\%) \\
    Iris          &   135,952 & 1,267 (-99.07\%) \\
    Tic-Tac-Toe   & 5,670,532 & 1,529 (-99.97\%) \\
    Vote          &   988,358 & 1,148 (-99.88\%) \\
    % Weather       &    85,860 &    93 (-99.89\%) \\
    \hline
  \end{tabular}
  \vspace{1em}
  \caption{Decision diagram sizes for Random Forests of size 10.000 for other datasets~\cite{uci-repo}.}
  \label{tab:size}
\end{table}
Breathing semantics into the decision diagrams by unsatisfiable path elimination overcomes the scalability problems that are due to the enormous space requirements. 
In fact, it avoids the exponential blow-up in size in all three variants, with DD* even becoming significantly smaller than the original Random Forest (cf. Fig.~\ref{fig:size}).
Moreover, also the classification times are drastically reduced in all three cases (cf. Fig.~\ref{tab:running-time}). 
In fact, the classification times eventually stabilise for DD*, illustrating the key feature of Random Forests, the reduction of the learner's variance. 
\footnote{Remember, being just a different representation of the original Random Forest, DD* has the same variance.} 
As sketched in Tables~\ref{tab:running-time} and~\ref{tab:size} these observations carry over to other popular data sets in the UCI Machine Learning Repository~\cite{uci-repo}. 

% \textbf{Parallelism for free either way:}
% Naively one could think that running time of a Random Forest can similarly be reduced by parallelisation of its evaluation.
% That is naturally possible by evaluating all its trees in parallel and aggregating the obtained result subsequently.
% Note however, that this potential is not lost:
% We can aggregate subsets of trees and produce an ensemble of decision diagrams rather than decision trees.
% In fact, we can choose the number of decision diagrams used to represent the Random Forest ideally matching the number of processing units available on the user machine.
% This way, we can benefit from both ideas simultaneously:
% We take full advantage of parallel processing while at the same time preventing redundant computation.
% 
\section{Conclusion}
\label{sec:conclusion}
In this paper, we have presented an approach to aggregate large Random Forests into a single and compact decision diagram using the machinery of Algebraic Decision Diagrams. 
This radical transformation allows for rapid classification and decision-making while, at the same time, preserving the original semantics. 
We have refined our method in multiple steps: by compositional abstraction, pre-evaluation at compile time, and the elimination of redundant predicates. 
As a result, we could achieve running time reductions by factors of thousands on multiple popular datasets, proving the method's relevance for real world problems. 
At the same time, and beyond our original expectations, also the size requirements could be reduced by orders of magnitude.

The results reported in this paper concern a widely used classifier and standard datasets and were achieved with clean aggregation, abstraction, and reduction criteria:
\begin{itemize}
  \item ADD-based aggregation is canonical as soon as an order of predicates has been fixed. 
  Thus the freedom of choice here reduces to the choice of an adequate variable ordering, a task heuristically taken care of by the corresponding frameworks~\cite{Somenzi2001}.
  \item \emph{Class frequency} abstraction is the coarsest, compositional abstraction that still allows one to faithfully represent the classification function of the original Random Forest.
  \item Unfeasible path elimination does not support normal forms, but the results are minimal, meaning that the resulting structures cannot be reduced further without changing the semantics of the classification function. 
  In essence, the variability here is a consequence of the freedom of choice where to \emph{root} unfeasible paths. 
  It can be seen as a generalisation of the classical problem of minimising Boolean functions with \emph{don't cares}.
  \item \emph{Most frequent class} abstraction reduces the final compositionally reduced decision diagrams to the smallest diagram that still represents the original classification function.
\end{itemize}
Thus our approach is optimal relative to two well-known conceptual hurdles, the choice of variable ordering for decision diagrams and the treatment of \emph{don't cares}. 
In particular, our approach does not exploit any peculiarities of certain classifiers or data sets. 

Of corse, its impact may still strongly depend on the structure of the concrete considered scenario.
We are therefore currently investigating how easily these results can be adopted to other data sets and classifiers. 
While we expect the transfer of these ideas to be relatively straight forward for some discrete classifiers, e.g. Decision Jungles~\cite{Shotton2013}, this will be more difficult for others. 
For more complex classifiers such as neural networks, compromises may be unavoidable. 
In these cases, rapid classification may come at the cost of semantic approximation. 

With this approach being so successful for the very specific domain of Random Forests, we are also interested in its generalisation ability. 
Rather than the optimisation of what can be considered a domain-specific program, it will be interesting to see its potential in the context of general-purpose programming languages.
First results have shown that we can indeed use very similar techniques to optimise more general programs~\cite{addCompiler}.

% Researchers in machine learning have already extracted rules from datasets and from classification models -- also from decision %tree ensembles~\cite{Hall1998,SzpunarHuk2006}.
% These rule-based approaches were motivated by their potential to reduce the ensemble complexity and to improve generalisation %capability.
% Interestingly, the resulting sets of rules allow for the application of miAamics~\cite{miaamics}, a domain-specific language for %decision function. 
% Together, these techniques could potentially resemble the results presented in this paper.
% More generally, this could lead the path for other radical transformations of a wide variety of classifiers. 
% 
\bibliographystyle{splncs04}
\bibliography{literature}
\end{document}

%% file: mathmacros.tex
\DeclareMathOperator*{\argmax}{arg\,max}